\pdfoutput=1

\documentclass[11pt]{article}
\usepackage{authblk}
\usepackage[final]{ACL2023}
\usepackage{times}
\usepackage{latexsym}
\usepackage{amssymb}
\usepackage{url}
\usepackage{booktabs}
\usepackage{graphicx}
\usepackage{float}
\usepackage{multirow}
\usepackage{xcolor}
\usepackage{colortbl}

\usepackage[T1]{fontenc}

\usepackage[utf8]{inputenc}

\usepackage{microtype}

\usepackage{inconsolata}

\title{User Embedding Model for Personalized Language Prompting}
\author{Sumanth Doddapaneni, Krishna Sayana\thanks{\ \ Correspondence to: \texttt{ksayana@google.com}}\ , Ambarish Jash, Sukhdeep Sodhi, Dima Kuzmin}
\affil[]{Google Research}
  
\definecolor{lightblue}{rgb}{0.93,0.95,1.0}
\definecolor{lightgreen}{rgb}{0.93,1.0,0.95}

\begin{document}
\maketitle
\begin{abstract}
Modeling long user histories plays a pivotal role in enhancing recommendation systems, allowing to capture users' evolving preferences, resulting in more precise and personalized recommendations. In this study, we tackle the challenges of modeling long user histories for preference understanding in natural language. Specifically, we introduce a new User Embedding Module (UEM) that efficiently processes user history in free-form text by compressing and representing them as embeddings, to use them as soft prompts to a LM. Our experiments demonstrate the superior capability of this approach in handling significantly longer histories compared to conventional text-based methods, yielding substantial improvements in predictive performance. Models trained using our approach exhibit substantial enhancements, with up to 0.21 and 0.25 F1 points improvement over the text-based prompting baselines. The main contribution of this research is to demonstrate the ability to bias language models via user signals. 
\end{abstract}

\section{Introduction}

\begin{figure}[h!]
    \centering
    \includegraphics[width=\columnwidth]{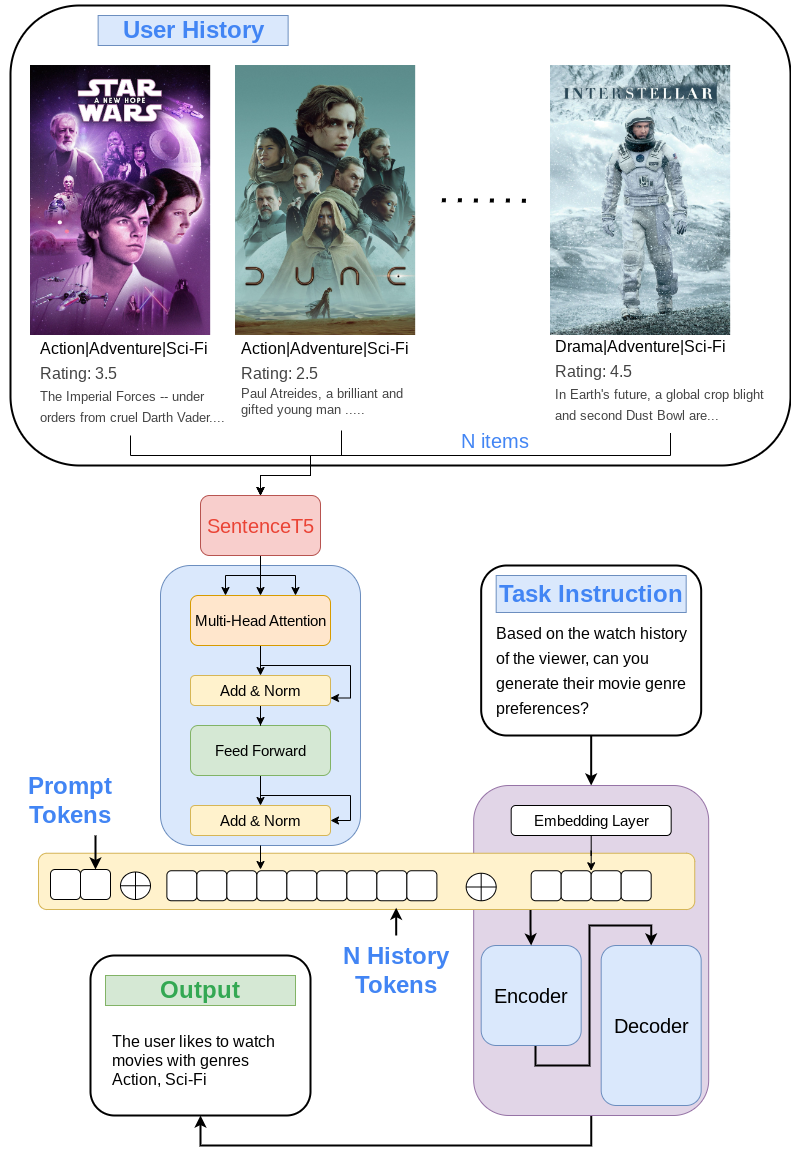}
    \caption{Overview of our User History Modeling Approach. The user history's textual features are processed through the User Embedding module and combined with the task prompt and subsequently passed through the language model.}
    \label{fig:teaser}
\end{figure}

In recent years, Large Language Models (LLMs) have proven their versatility in various language tasks, from translation to reasoning \cite{DBLP:journals/corr/abs-2303-12712}. Scaling up models and data has played a crucial role in unlocking their potential \cite{gpt4, palm2, llama2}. LLMs have also been adapted for conversational tasks, instruction following, and reasoning using techniques like Instruction-Tuning \cite{DBLP:conf/acl/MishraKBH22, flan, t0}, RLHF \cite{instructgpt}, and Chain-of-Thought \cite{CoT}. 
Trained on extensive internet data, these LLMs excel in generalization. They can quickly adapt to new tasks with in-context learning and are capable of not only answering questions but also reasoning about their responses.

The usage of LLMs has evolved beyond traditional NLP tasks to encompass tasks demanding reasoning \cite{qiao-etal-2023-reasoning}, long-form generation \cite{instructgpt}, creativity \cite{DBLP:journals/corr/abs-2305-05576}, and demonstrated remarkable proficiency in these areas. LLMs have been applied to search, retrieval, ranking, chat, personalization, recommendation systems, and others \cite{DBLP:conf/icml/YasunagaAS0LLLZ23, instructgpt, lamp}. One practical use case for LMs is understanding \textit{user preferences} to \textit{generate recommendations}, a task that extends beyond text to encompass audio and visual modalities in real-world scenarios, as exemplified by platforms like YouTube\footnote{\url{youtube.com}}, Spotify\footnote{\url{spotify.com}} among many others.

Recent research has predominantly concentrated on examining smaller segments of user history by selecting representative samples from a users' history \cite{lamp}. \citet{mu2023learning} uses learned gist tokens to compress prompts while \citet{li2023automatic} uses prompt rewriting based on entries retrieved from a users' profile. This leads to the critical question \textit{How can we effectively utilize longer user histories?} To achieve this, we employ an embedding-based technique to compress the user's entire history, 
creating a sequence of representative user embedding tokens. This embedded representation enhances our ability to comprehend user preferences and subsequently generate predictions that align more closely with their interests. Further, since the User Embedding Module (UEM) module is co-trained with the LM, the representations are learned in-context for the specific tasks. Our research demonstrates the advantages of this approach, particularly in its capacity to incorporate longer user history into LMs, resulting in more robust user preference understanding. Compared to the naive approach of concatenating user history and incurring $O(n^2)$ compute cost for self-attention, our approach demonstrates a cheap way to incorporate history metadata as an embedding thus dramatically reducing the required compute. As a result longer user histories can be easily incorporated within LMs. Our empirical findings demonstrate the ability of our approach to accommodate significantly larger histories compared to traditional text-based methods, resulting in improved predictive performance.

\section{Approach}
\label{sec:approach}

Following text-to-text approach of T5 \cite{t5}, we frame all tasks as text generation conditioned on the input. Formally, given a sequence of query input tokens denoted as X, we model the probability of output Y as $Pr_{\theta}(Y|X)$, where $\theta$ represents the weights of the model. Prior studies have established two primary prompting strategies: \textit{text-based prompting}, where textual instructions are prepended to the input \cite{DBLP:conf/acl/MishraKBH22, flant5, CoT}, and \textit{soft-prompting}, which adds a set of trainable tokens as a prefix to the input tokens of the models \cite{lester-etal-2021-power, prefix-tuning}. 
Prior soft-prompting uses a fixed task-specific soft-prompt to achieve parameter-efficient fine-tuning for various language tasks, maximizing the likelihood $Pr_{\theta}(Y|[K; X])$, with K trainable tokens. We extend this idea to personalization. More specifically, using the User Embedding Module (UEM), we generate a personalised soft-prompt conditioned on the users' history. This setup aims to maximize the likelihood of the label Y, given $Pr_{\theta}(Y|[Pr_{UEM}(U); X])$, where $Pr_{UEM}(U)$ are the soft prompts generated by the UEM based on user history $U$ and prefixed to the query input $X$. In our task definition, $U$ corresponds to the movie metadata, $X$ represents the task instruction, and $Y$ represents genre preferences.

Overall, given a task instruction $X$, the LM embeds these tokens to a matrix $X_e \in \mathbb{R}^{n \times e}$, where \textit{n} represents the token count and \textit{e} represents the embedding dimension of the LM. The textual user history $H = \{h_i\}_{i=1}^p$ is converted into embeddings $U = \{u_i\}_{i=1}^p$ with SentenceT5 \cite{sentencet5}. Each history item $u_i$ is a composite of three distinct embeddings: (i) title \& genre, (ii) rating, and (iii) description. The collective history of `p' items is expressed as $U \in \mathbb{R}^{p \times 3s}$, where \textit{s} corresponds to the embedding dimension of SentenceT5. These embeddings undergo processing within a transformer network (UEM). To ensure dimension alignment with \textit{e}, a linear projection layer is introduced atop the transformer, mapping the dimension \textit{3s} to \textit{e}, thereby yielding $Pr_{UEM}(U) \in \mathbb{R}^{p \times e}$.

Following \citet{lester-etal-2021-power}, we also incorporate `k' task-level soft prompts, denoted as $P_e \in \mathbb{R}^{k \times e}$. Both the user and task prompts are concatenated with the input embedding, resulting in a unified embedding matrix, represented as $[P_e; Pr_{UEM}(U); X_e] \in \mathbb{R}^{(k+p+n) \times e}$. This composite embedding flows through the LM, maximizing the probability of Y, and concurrently updating all parameters within both model components. The model is illustrated in Figure \ref{fig:teaser}.

\section{Experiments}

\paragraph{Implementation.} As described in \S\ref{sec:dataset}, we use the MovieLens dataset \cite{movielens} in conjunction with movie descriptions. For the embeddings $U$ discussed in \S\ref{sec:approach}, we format the text in the following manner: (i) title and genre - \texttt{The movie \{movie\_title\} is listed with genres \{genres\}}, (ii) rating - \texttt{The movie is rated with \{rating\} stars}, and (iii) description - \texttt{\{movie\_description\}}. In the case of the text-only baselines, we input the concatenated strings instead of the embeddings. The dataset i split into 117k/5k/5k for train, validation and test sets respectively. Unless specified, we use the FlanT5 \cite{flant5} series of models for all experiments, training them for 10k steps with a batch size of 128. Text-history models use a learning rate of 1e-2, while embedding-history models use 5e-3. Our user embedding model consists of 3 transformer layers with 12 attention heads, 768d embeddings, and 2048d MLP layers, adding 65M parameters. We use 20 tokens for task-level soft prompts $k$.

\paragraph{Evaluation.} Although the task is framed in a text-to-text format, the model's output can be processed by a verbalizer to extract the genres. While conventional metrics like BLEU \cite{bleu}, ROUGE \cite{rouge}, and COMET \cite{comet} are used for evaluating generative text, they lack granularity in understanding the task performance. However, given the straightforward genre extraction by the verbalizer, we treat the task as a multi-label classification problem and present weighted precision, recall and F1 scores across all labels\footnote{There are 19 genres with a high skew among the classes. We use \texttt{sklearn.metrics.classification\_report}}. Our initial findings indicate that these scores offer a more interpretable assessment, both at the genre level and for the overall task evaluation, compared to token-level metrics. 

\subsection{Main Results}
We present the results from our proposed approach in Table \ref{tab:main}. The results demonstrate that incorporating a larger history significantly enhances the models' understanding of the user preferences. 
Compared to the text-only models, we observe F1 improvements of 0.21 and 0.25 in performance for the base and large models, respectively. To assess performance against text-only models with a comparable history size, we train a model with only 5 history items. The results reveal slightly poorer performance, likely due to the extremely limited context window of the history (5 tokens) compared to the text-only model (over 1000 tokens). 
Unlike conventional language models, LongT5 \cite{longt5}, is trained with Transient Global Attention, allowing it to efficiently process longer text sequences. However, it's essential to consider that this extended capability comes at the cost of increased memory and longer training times. While FlanT5 models can be effectively trained on v3-8 TPUs, LongT5 necessitates v3-32 TPUs and requires 4x the training time of a FlanT5 model of comparable size, especially when dealing with input sequences of 16k tokens (equivalent to 50 history items). More importantly, the serving latency is also correspondingly increased, which could make these models impractical for production use.

\begin{table}[h!]
\small
\centering
\begin{tabular}{llcc}
\toprule
 &  & \texttt{BASE} & \texttt{LARGE} \\
 \midrule
\multirow{3}{*}{\begin{tabular}[c]{@{}l@{}}Counting\\ Baselines\end{tabular}} & precision & \multicolumn{2}{c}{0.330} \\
 & recall & \multicolumn{2}{c}{0.273} \\
 & f1 & \multicolumn{2}{c}{0.192} \\
 \cmidrule(lr){1-4}
 \rowcolor{lightgreen}  & precision & 0.276 & 0.257 \\
 \rowcolor{lightgreen}  & recall & 0.287 & 0.273 \\
 \rowcolor{lightgreen} \multirow{-3}{*}{Text Hist. \texttt{5}} & f1 & 0.273 & 0.261 \\
 \cmidrule(lr){1-4}
 \rowcolor{lightgreen}  & precision & 0.275 & 0.281 \\
 \rowcolor{lightgreen}  & recall & 0.297 & 0.290 \\
 \rowcolor{lightgreen} \multirow{-3}{*}{Emb. Hist. \texttt{5}} & f1 & 0.252 & 0.215 \\
 \cmidrule(lr){1-4}
 \rowcolor{lightblue} & precision & 0.541 & 0.568 \\
 \rowcolor{lightblue}  & recall & 0.523 & 0.558 \\
 \rowcolor{lightblue} \multirow{-3}{*}{LongT5 \texttt{50}} & f1 & 0.529 & 0.557 \\
 \cmidrule(lr){1-4}
 \rowcolor{lightblue}  & precision & 0.407 & 0.400 \\
 \rowcolor{lightblue}  & recall & 0.405 & 0.399 \\
 \rowcolor{lightblue} \multirow{-3}{*}{Emb. Hist. \texttt{50}} & f1 & 0.396 & 0.381 \\
 \cmidrule(lr){1-4}
\multirow{3}{*}{Emb. Hist. \texttt{100}} & precision & 0.416 & 0.459 \\
 & recall & 0.413 & 0.441 \\
 & f1 & 0.404 & 0.444 \\
 \bottomrule
\end{tabular}
\caption{Model performance using proposed User Embedding Module. \textit{Counting Baselines} refers to counting the three most frequently occurring genres across the entire user history.}
\label{tab:main}
\end{table}

\subsection{Ablations}
\begin{figure}[h]
    \centering
    \includegraphics[width=\columnwidth]{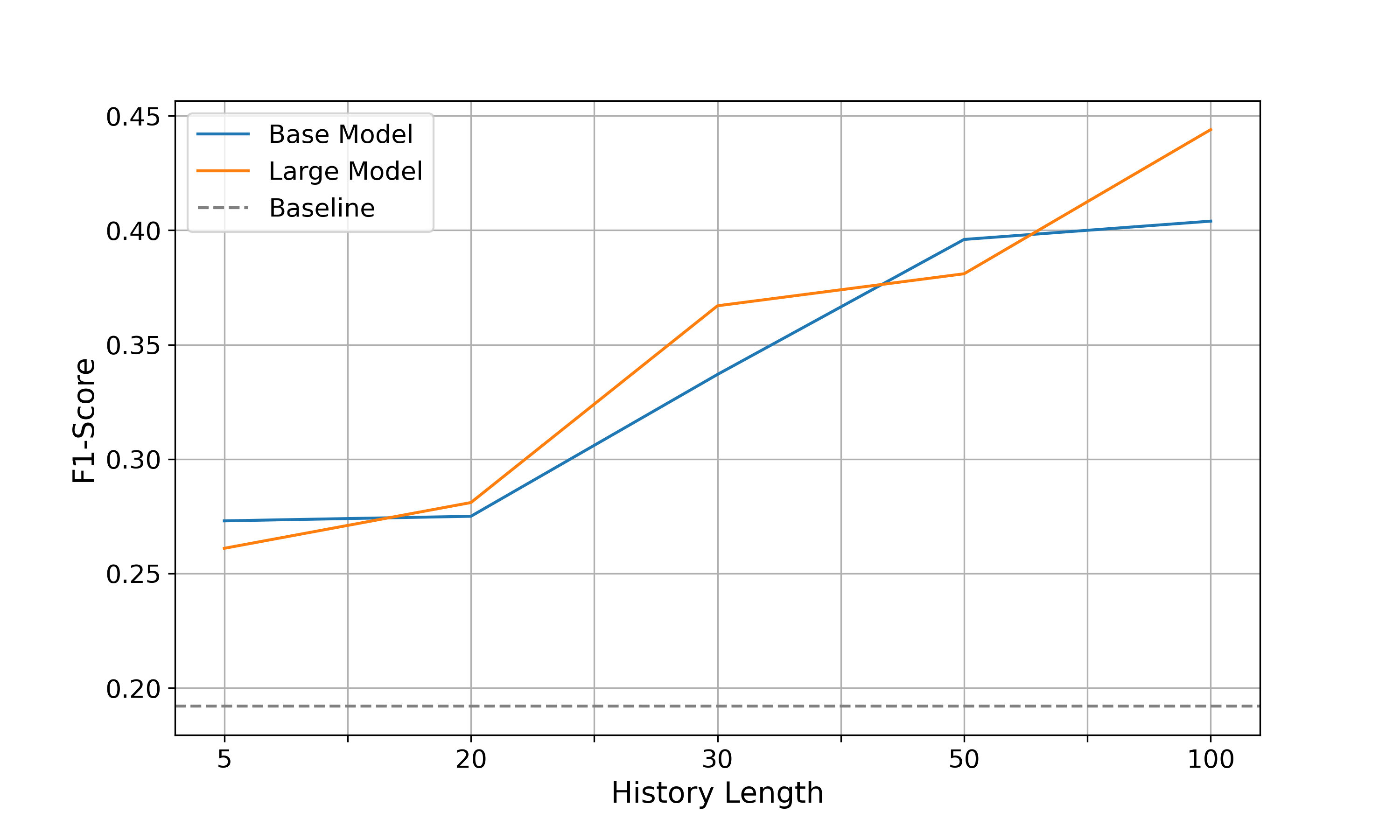}
    \caption{Comparison of model performance with increasing User History.}
    \label{fig:hist-len}
\end{figure}

\paragraph{Effect of History Length.}To assess the impact of the history size, we conduct a series of ablations by increasing the users' history passed to UEM. The results are presented in Figure \ref{fig:hist-len} (ref. Table \ref{tab:hist-len}), revealing an improvement in model performance with an increase in the number of history items. It's worth noting that incorporating 50 history items in textual form results in an input of nearly 16k tokens. While methods like LongT5 \cite{longt5}, ALiBi \cite{alibi}, and ROPE \cite{rope} allow for extrapolation to longer sequences, this remains computationally intensive. 

\paragraph{Choice of LM.}In our experiments, we chose FlanT5 as the language model for our task. We also conducted experiments with both the base T51.1 \cite{t5} and a LM adapted T5 model \cite{lester-etal-2021-power} to find the best starting point for our training. The results presented in Table \ref{tab:lm-choice} shows that FlanT5 has the best performance, mainly due to the instructional nature of our task prompts. 

\begin{table}[h!]
\centering
\small
\begin{tabular}{llcc}
\toprule
 &  & \texttt{BASE} & \texttt{LARGE} \\
 \midrule
\multirow{3}{*}{T5$_{1.1}$} & precision & 0.282 & 0.322 \\
 & recall & 0.292 & 0.324 \\
 & f1 & 0.208 & 0.267 \\
 \cmidrule(lr){1-4}
\multirow{3}{*}{T5$_{LM Adapted}$} & precision & 0.353 & 0.398 \\
 & recall & 0.374 & 0.397 \\
 & f1 & 0.338 & 0.378 \\
 \cmidrule(lr){1-4}
\multirow{3}{*}{FlanT5} & precision & 0.407 & 0.400 \\
 & recall & 0.405 & 0.399 \\
 & f1 & 0.396 & 0.381 \\
 \bottomrule
\end{tabular}
\caption{Comparison of model performance with various choices of Language Models.}
\label{tab:lm-choice}
\end{table}

\paragraph{Size of UEM.}For the user embedding module, we experimented with different sizes by changing the number of layers in the transformer block. We found that gradually making UEM bigger improved the performance of both the base and large models (see Table \ref{tab:uem-size}). However, we also recognize that the task itself may not require a very complex solution, so further increasing the size of the module may not be justified. We believe that different tasks might need different levels of complexity, and we plan to explore this in future research.

\begin{table}[h!]
\centering
\small
\begin{tabular}{llcc}
\toprule
 &  & \texttt{BASE} & \texttt{LARGE} \\
 \midrule
\multirow{3}{*}{1 Layer} & precision & 0.391 & 0.395 \\
 & recall & 0.381 & 0.384 \\
 & f1 & 0.346 & 0.347 \\
 \cmidrule(lr){1-4}
\multirow{3}{*}{2 Layers} & precision & 0.399 & 0.380 \\
 & recall & 0.392 & 0.367 \\
 & f1 & 0.384 & 0.365 \\
 \cmidrule(lr){1-4}
\multirow{3}{*}{3 Layers} & precision & 0.407 & \multicolumn{1}{r}{0.400} \\
 & recall & 0.405 & \multicolumn{1}{r}{0.399} \\
 & f1 & 0.396 & \multicolumn{1}{r}{0.381} \\
 \bottomrule
\end{tabular}
\caption{Comparison of model performance with various sizes of User Embedding Module. All the models use the same history size of 50.}
\label{tab:uem-size}
\end{table}

\section{Related Work}

In prior research, UserAdapter \cite{useradapter} introduced a trainable token for each user, facilitating sentiment classification specialization using RoBERTa. Expanding upon this, UserIdentifier \cite{useridentifier} demonstrated that employing random userIDs effectively captures user-specific information. HuLM \cite{hulm} pretrained a LM conditioned on higher-order data states associated with humans. Further, \citet{lamp} employed retrievers like Contriver and BM25 to select representative input histories to prompt an LM to generate personalized outputs. \citet{mu2023learning} utilized gist tokens to condense input prompts into a set of tokens, reducing computational overhead for recurring task instructions. \citet{li2023automatic} employed prompt rewriting, identifying relevant items for individual users, summarizing the information, and synthesizing key attributes to prompt the model. Our approach distinguishes itself by utilizing entire user histories, compressing them into contextually learned embeddings.

\section{Conclusion \& Future Work}
In this study, we addressed several critical challenges in modeling user history for preference understanding. We introduced a User Embedding Module that processed user history as freeform text, generating token embeddings for each history item. This approach greatly simplified user history tracking and enabled the incorporation of longer user histories into the language model, and allowed their representations to be learned in context. Our empirical results demonstrated the capability of this approach to handle significantly larger histories efficiently compared to traditional text-based approaches, resulting in improved predictive performance. For future work, we would like to explore more parameter efficient approaches like LoRA \cite{hu2021lora} for finetuning LMs with UEM, which would improve both training and serving for these models. This approach can be easily extended to multimodal signals using modal specific embeddings and tying them together with UEM, and we plan to explore this direction for future work.

\section*{Limitations}
While we argue and demonstrate in this work that using a UEM is an efficient way to encode long user histories with easier extensions to multimodal inputs, we acknowledge that text prompting can be further optimized, by using text-to-text prompt compression models. These trade-offs could be further studied. The simplicity of the UEM architecture leaves a lot of headroom as demonstrated by LongT5 baselines in Table \ref{tab:main}. Our presentations for $U$ are 
using generic semantic embeddings with the use of SentenceT5 \cite{sentencet5}, these can be further improved with the use of domain specific embeddings. Our experiments are using LMs that are <1B parameters, which are usually considered smaller family of LLMs. It would be a good future direction to consider larger models with parameter efficient tuning techniques. Furthermore, our research has primarily focused on preference understanding, and hasn't been tested on tasks extending to areas such as rating prediction or item recommendation. We expect our conclusions here are likely apply to these tasks. We plan to address these limitations and pursue these avenues in our future research efforts.

\section*{Ethics Statement}
The datasets and models utilized in this study are based on publicly available and open-source resources. While we acknowledge the inherent ethical considerations associated with language models, we do not anticipate any additional ethical concerns arising from the datasets and models developed in the course of this research.

\bibliography{anthology,custom}
\bibliographystyle{acl_natbib}

\appendix

\section{Dataset}
\label{sec:dataset}
In the MovieLens dataset \cite{movielens}, the available metadata for assessing a movie's rating is confined to its title and associated genres. However, such limited information proves inadequate for both human evaluators and language models in generating predictions unless they possess prior knowledge about the movies. For instance, when considering the \textit{Star Wars} series within the MovieLens dataset \cite{movielens}, namely \textit{Star Wars: Episode IV - A New Hope (1977)}, \textit{Star Wars: Episode VI - Return of the Jedi (1983)}, and \textit{Star Wars: Episode I - The Phantom Menace (1999)}, all three films share identical genre classifications, namely Action, Adventure, and Sci-Fi. Nonetheless, a closer inspection reveals noteworthy disparities in their mean ratings, with "Episode IV" and "Episode VI" accumulating ratings of \texttt{4.12} and \texttt{4.14}, respectively, while "Episode I" registers a markedly lower rating of \texttt{3.06}.

In reality, a viewer's decision to watch a movie is contingent upon a multifaceted array of metadata beyond the movie genres. Variables such as the cast, crew, production studio, among others, play pivotal roles in this determination. In contemporary times, movie trailers have emerged as potent tools for piquing an individual's interest in a movie. In the context of textual language models, we equate the concept of a \textit{gist} as a close analogue to a movie trailer. The gist encapsulates the fundamental essence of the movie's content while withholding explicit plot details, a characteristic akin to that of a trailer. Consequently, we propose the incorporation of such supplementary data\footnote{Metadata sourced from \url{https://www.kaggle.com/datasets/stefanoleone992/rotten-tomatoes-movies-and-critic-reviews-dataset/}} into the MovieLens dataset \cite{movielens} to facilitate more nuanced and informed predictive assessments. While it is acknowledged that this augmentation may not encompass the entirety of a viewer's decision-making process, it represents a stride closer to the intricacies involved in real-world movie-watching choices.

After merging the MovieLens dataset \cite{movielens} with movie descriptions and filtering out users with fewer than 20 recorded movie views, our dataset comprises 14.4M reviews, spanning 8.2k unique movies, and involving a total of 127k users. We then divide this dataset into three subsets: 5k users for both the development and testing sets, and the remaining 117k users for the training set. To create gold labels, we aggregate genres along with their corresponding ratings across each user's viewing history. Only genres with a minimum of three ratings are considered. Based on this aggregated information, we identify the three most preferred genres (with an average rating >3.5) and the three least preferred genres (with an average rating <3) for each user. The resulting output is structured in a text-to-text format as follows:  \texttt{The user likes to watch movies with genres \{liked\_genres\} and doesn\textquotesingle t like to watch movies with genres \{disliked\_genres\}}\footnote{In cases where the set of \texttt{liked\_genres} or \texttt{disliked\_genres} is empty, the text is adjusted accordingly.}.

\section{History Length Ablation Results}
\label{apx:hist-len}

\begin{table}[h!]
\centering
\small
\begin{tabular}{llrr}
\toprule
 &  & \texttt{BASE} & \texttt{LARGE} \\
 \midrule
\multirow{3}{*}{Emb. Hist. \texttt{5}} & precision & 0.276 & 0.257 \\
 & recall & 0.287 & 0.273 \\
 & f1 & 0.273 & 0.261 \\
 \cmidrule(lr){1-4}
\multirow{3}{*}{Emb. Hist. \texttt{20}} & precision & 0.319 & 0.321 \\
 & recall & 0.328 & 0.326 \\
 & f1 & 0.275 & 0.281 \\
 \cmidrule(lr){1-4}
\multirow{3}{*}{Emb. Hist. \texttt{30}} & precision & 0.353 & 0.390 \\
 & recall & 0.364 & 0.390 \\
 & f1 & 0.337 & 0.367 \\
 \cmidrule(lr){1-4}
\multirow{3}{*}{Emb. Hist. \texttt{50}} & precision & 0.407 & 0.400 \\
 & recall & 0.405 & 0.399 \\
 & f1 & 0.396 & 0.381 \\
 \cmidrule(lr){1-4}
\multirow{3}{*}{Emb. Hist. \texttt{100}} & precision & 0.416 & 0.459 \\
 & recall & 0.413 & 0.441 \\
 & f1 & 0.404 & 0.444 \\
 \bottomrule
\end{tabular}
\caption{Comparison of model performance with increasing User History.}
\label{tab:hist-len}
\end{table}


\end{document}